\def\year{2020}\relax
\documentclass[letterpaper]{article} 
\usepackage{aaai20}  
\usepackage{times}  
\usepackage{helvet} 
\usepackage{courier}  
\usepackage[hyphens]{url}  
\usepackage{graphicx} 
\usepackage{graphicx}  
\usepackage[misc]{ifsym}
\usepackage{amsfonts}
\usepackage{amsmath}
\usepackage[ruled,vlined,linesnumbered]{algorithm2e}
\usepackage{array}
\usepackage{multirow}
\usepackage{subcaption}
\usepackage{dblfloatfix}

\newcolumntype{L}[1]{>{\raggedright\let\newline\\\arraybackslash\hspace{0pt}}m{#1}}
\newcolumntype{C}[1]{>{\centering\let\newline\\\arraybackslash\hspace{0pt}}m{#1}}
\newcolumntype{R}[1]{>{\raggedleft\let\newline\\\arraybackslash\hspace{0pt}}m{#1}}

\newcommand{\argmax}{\operatornamewithlimits{argmax}}
\newcommand{\colwidth}{}
\newcommand{\citet}[1]{\citeauthor{#1}~\shortcite{#1}}

\title{Identifying Model Weakness with Adversarial Examiner}
\author{Michelle Shu, Chenxi Liu\textsuperscript{(\Letter)}, Weichao Qiu, Alan Yuille \\ 
Johns Hopkins University\\ 
mshu1@jhu.edu, cxliu@jhu.edu, qiuwch@gmail.com, alan.l.yuille@gmail.com 
}
 
\begin{document}

\maketitle

\begin{abstract}
Machine learning models are usually evaluated according to the average case performance on the test set. However, this is not always ideal, because in some sensitive domains (e.g. autonomous driving), it is the worst case performance that matters more. In this paper, we are interested in systematic exploration of the input data space to identify the weakness of the model to be evaluated. We propose to use an \emph{adversarial examiner} in the testing stage. Different from the existing strategy to always give the same (distribution of) test data, the adversarial examiner will dynamically select the next test data to hand out based on the testing history so far, with the goal being to undermine the model's performance. This sequence of test data not only helps us understand the current model, but also serves as constructive feedback to help improve the model in the next iteration. We conduct experiments on ShapeNet object classification. We show that our adversarial examiner can successfully put more emphasis on the weakness of the model, preventing performance estimates from being overly optimistic. 
\end{abstract}

\section{Introduction}

The field of machine learning is advancing at unprecedented speed, and one evidence often cited is the reported performance on benchmark datasets. 
There are usually leaderboards for these public datasets with specific evaluation metrics, and the entry that achieves the highest number is regarded as the state-of-the-art. 
In several cases, it is claimed that the machine learning model has surpassed human-level performance using this criterion \cite{he2015delving,wang2018multi}.

However, the common belief is that humans are still superior, which suggests that the current testing strategy is overly optimistic and does not reflect the true progress in advancing machine learning. 
We think an important reason behind this mismatch is that: the current evaluation protocol uses \emph{fixed} test data and measures the \emph{average} case performance, which does not place enough emphasis on the \emph{worst} case performance. 
To make this point concrete, imagine a model for autonomous driving. 
Suppose that in the testing data, there are 1 million images under normal conditions (in which case the car should keep driving), and 1 image with a baby in front (in which case the car should immediately stop). 
Under the current evaluation protocol, a policy to always keep driving and give up on the rare case will be highly regarded. 
However, from the humans' perspective, this strategy is clearly problematic and alarming. 
Therefore, at least in some sensitive domains, placing more emphasis on the worst cases to reveal model weaknesses may be more important than simply relying on the average case. 

Revealing model weaknesses by densely sampling the input data space is clearly impractical. 
This is because any input data point is usually determined by a combination of factors, which makes the space exponentially large. 
For example, an image is jointly determined by the rotation angles of the objects, the positions of the objects, the material properties, the type of lighting, the energy of lighting, the distance to camera, etc. 
Larger test set certainly helps, but no matter how large it gets, it will never be able to densely cover this exponentially large space. 
Therefore our goal is to efficiently and systematically identify the model weaknesses within a reasonable number of queries. 

\begin{figure*}[t]
    \centering
    \includegraphics[width=0.8\textwidth]{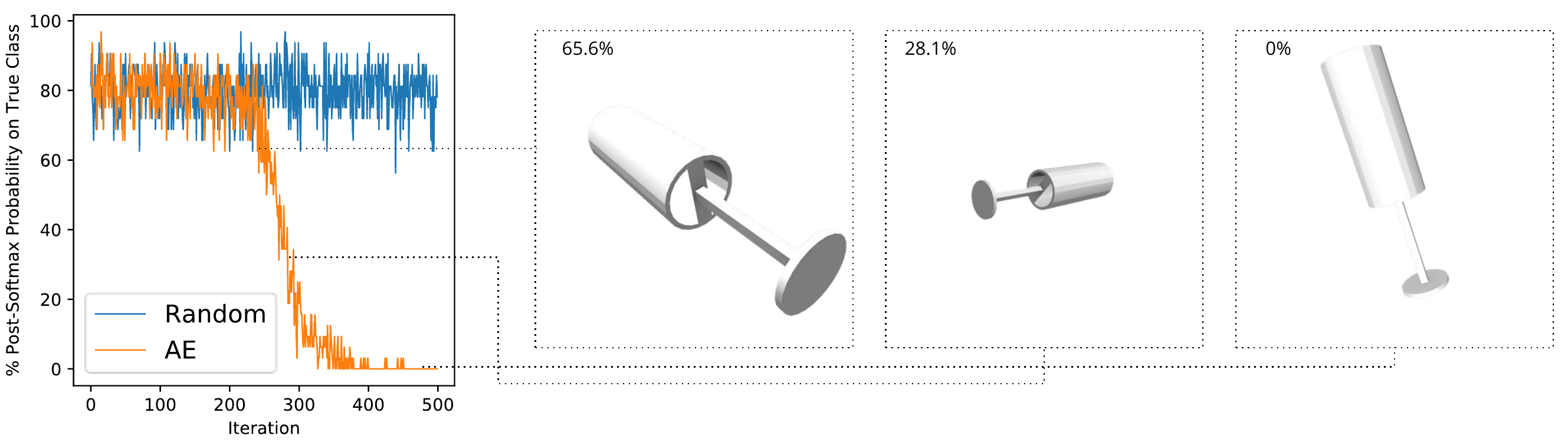}
    \caption{Evaluating a model's ability to recognize a \textit{lamp} instance in ShapeNet. 
    If we randomly sample data, the evaluation will converge to a high percentage (around $80\%$).
    However, using our Adversarial Examiner (AE), the test cases gradually concentrate on the weaknesses of the model, and the evaluation will not be overly optimistic.
    In this example, AE has found that the model appears to be vulnerable to increased lighting (within a reasonable range). }
    \label{fig:examine_examples}
\end{figure*}

Different models have different weaknesses, so it no longer makes sense to have a fixed test set, rather a dynamic one.
In some way, this notion of fixed versus dynamic is very similar to giving standardized tests versus conducting interviews. 
In standardized tests, the questions are usually designed to have general coverage, since all test takers answer the same questions. 
In interviews, the examiner may start with general topics, but also has the freedom to deep dive into specific ones to challenge the candidate. 
We also point out that this notion of dynamic examiner is closer in spirit to Turing test.
If the human evaluator in Turing test always asks the same questions, then its role can be replaced by a machine. 
It is also conceivable that the worst answer during the exchange will contribute most to the human evaluator's final judgment. 

Formalizing the ideas discussed above, we propose \emph{adversarial examiner}, an alternative way of evaluating a model.
Provided with a model and the space of input data, the adversarial examiner will dynamically select the next test data to hand out, based on the testing history so far. 
The name ``adversarial'' comes from the fact that its goal is to minimize the model's performance. 
Compared with the standard testing protocol, adversarial examiner obviously places more emphasis on the worst case scenarios. 
Because of its dependence on testing history, the sequence of test data naturally reveals the process of identifying model weakness. 
Fig.~\ref{fig:examine_examples} offers a demonstration. 

Adversarial attack is also a way of showing the weakness of the model. 
The message of adversarial attack is that for a particular data point, there exists another data point within its local neighborhood, such that the model fails. 
However, both the original data point and the local neighborhood are microscopic properties. 
By contrast, our adversarial examiner starts with the entire input data space, and therefore has the potential of revealing the global picture. 
In addition, since the adversarial examiner concerns the global structure, the testing results are more informative and more easily perceived as ``constructive feedback'' that help improve the model in the next iteration. 

We conduct experiments on ShapeNet object classification. 
The input data space is the Cartesian product of $12$ physical parameters that determine the scene, including the rotation and position of object, the energy and angle of lighting, etc.
Through both quantitative evaluation and qualitative visualization, we show that our adversarial examiner can successfully identify the weakness of the model, preventing performance estimates from being overly optimistic. 

\section{Method}

\subsection{Adversarial Examiner}

We begin by revisiting the evaluation protocol employed in a standard classification task.
The examples $x \in \mathbb{R}^p$ follow an underlying distribution $\mathcal{P}$. 
The ground truth label of $x$ is $y(x) \in \{1, 2, \hdots, k\}$. 
The label predicted by the model $f$ is $f(x) \in \mathbb{R}^k$. 
Given a loss function $L(\cdot, \cdot)$ that measures the compatibility between $f(x)$ and $y(x)$, the evaluation metric is:
\begin{equation}
    E = \mathbb{E}_{x \sim \mathcal{P}} [L(f(x), y(x))] \approx \frac{1}{N} \sum_{i=1}^N L(f(x_i), y(x_i))
    \label{eqn:std}
\end{equation}
where $x_i \sim \mathcal{P}$. 

In applications such as computer vision and natural language processing, the examples $x$ are usually the surface representation, e.g. 2D images in vision and words in language. 
However, there is typically an underlying representation $z$ that generates/determines this surface representation $x$. 
In vision, $z$ would be the 3D object instance.
In language, $z$ would be the semantics/pragmatics that the speaker wishes to communicate. 
In the following, we consider evaluation on the underlying representation $z$, which is more intrinsic.

The examples $z \in \mathbb{R}^q$ follow an underlying distribution $\mathcal{Q}$, whose support is $\mathcal{D}$.
If we can obtain the ground truth label using the surface form $x$, then we can certainly obtain it using the underlying form $z$. 
We slightly abuse annotation and use $y(z) \in \{1, 2, \hdots, k\}$ to represent the ground truth label for instance $z$. 
Let $g$ be the function that transforms the underlying form into surface form $x = g(z, s)$, where $s \in \mathcal{S}$ represents the remaining information needed to complete this transformation. 
For example, in vision, $s$ could include camera information, lighting condition, etc.
In language, $s$ may be a tie-breaker between synonyms or sentence templates that express the same meaning. 

We use the following evaluation metric in our \emph{adversarial examiner}:
\begin{equation}
    \begin{split}
    E_{\text{examiner}} &= \mathbb{E}_{z \sim \mathcal{Q}} [\max_{s \in \mathcal{S}} L(f(g(z, s)), y(z))]\\ & \approx \frac{1}{N} \sum_{i=1}^N \max_{s_i \in \mathcal{S}} L(f(g(z_i, s_i)), y(z_i))
    \label{eqn:ae}
    \end{split}
\end{equation}
where $z_i \sim \mathcal{Q}$.
We choose $\mathcal{S}$ such that for any instance $z \in \mathcal{D}$ and any $s \in \mathcal{S}$, the surface form $g(z, s)$ still preserves the label $y(z)$ as judged by humans. 
The $\max$ in (\ref{eqn:ae}) means the examiner will consider the worst case within the space $\mathcal{S}$, which differs from (\ref{eqn:std}). 

It is easy to see that
\begin{equation}
    E_{\text{examiner}} \leq \max_{z \in \mathcal{D}, s \in \mathcal{S}} L(f(g(z, s)), y(z)) = L(f(g(\hat{z}, \hat{s})), y(\hat{z}))
\end{equation}
where $(\hat{z}, \hat{s})$ is the worst case scenario. 
We now discuss a special case, where the space $\mathcal{S}$ is so rich, that $\forall z_i, z_j \in \mathcal{D}$, $\exists s_i, s_j \in \mathcal{S}$, such that $g(z_i, s_i) = g(z_j, s_j)$. 
In this case, the worst combination $(\hat{z}, \hat{s})$ can always be achieved from any sample $z_i$. 
Therefore, $E_{\text{examiner}}$ will return \emph{exactly} the worst case performance under model $f$, which is what we expect and desire. 

Implementing the adversarial examiner according to (\ref{eqn:ae}) requires solving an optimization problem for every instance $z_i$. 
Since the landscape of $f(g(z_i, \cdot))$ on $\mathcal{S}$ can be arbitrary, we address the general case by assuming no closed-form solution and no differentiability. 
We consider a sequential strategy, where the adversarial examiner will hand out new candidates of $s_i^t$ based on the testing history so far $s_i^1, s_i^2, \hdots, s_i^{t-1}$. 
See Algorithm~\ref{alg:ae} for the adversarial examiner procedure. 

\begin{algorithm}[t]
\KwIn{$N$ samples $z_i \sim \mathcal{Q}$ and their true labels $y(z_i)$; Maximum number of examination steps $T$; Loss function $L$; Model $f$; Function $g$; Space $\mathcal{S}$.}
 \For{$i = 1$ \KwTo $N$}{
  Initialize \texttt{examiner} with $\mathcal{S}$\\
  \For{$t = 1$ \KwTo $T$}{
    $s_i^t =$ \texttt{examiner.generate()}\\
    $l_i^t = L(f(g(z_i, s_i^t)), y(z_i))$\\
    \texttt{examiner.update($s_i^t, l_i^t$)}\
  }
 }
 \Return{$E_{\text{examiner}} = \frac{1}{N} \sum_{i=1}^N l_i^T$}
\caption{Adversarial Examiner Procedure}
\label{alg:ae}
\end{algorithm}

\renewcommand{\colwidth}{0.6cm}

\begin{table*}[b]
    \centering
\begin{tabular}{|C{\colwidth}|C{\colwidth}|C{\colwidth}|C{\colwidth}|C{\colwidth}|C{\colwidth}|C{\colwidth}|C{\colwidth}|C{\colwidth}|C{\colwidth}|C{\colwidth}|C{\colwidth}|C{\colwidth}|}
\hline
   & $\alpha_{o}$ & $\beta_{o}$ & $\zeta_{o}$ & $\Gamma_{o}$ & $\Gamma_{l}$ & $r_{l}$ & $A_{l}$ & $U_{l}$  & $r_{v}$ & $A_{v}$ & $U_{v}$ & $\theta_{v}$ \\
\hline\hline
UB &$2\pi$ & $2\pi$ & $2\pi$ & 5 & 1 & 20 & 360 & 90 & 5 & 180 & 90 & 360     \\
\hline
LB & 0 & 0 & 0 & 0& 0.3 & 8 & 0 & -90 & 1 & 0 & -90 & 0     \\
\hline
\end{tabular}
    \caption{Upper bound (UB) and lower bound (LB) of rendering factors in $s$: sun rotation angles ($\alpha_o, \beta_o, \zeta_o$), sun energy ($\Gamma_o$), point light energy ($\Gamma_l$), point light distance ($r_l$), point light location ($A_{l}, U_{l}$), viewpoint distance ($r_v$), viewpoint location ($A_{v}, U_{v}$), viewpoint angle ($\theta_{v}$).}
    \label{tab:factors}
\end{table*}

\paragraph{Relation to Adversarial Attacks}

We named our examiner ``adversarial'' to acknowledge its connection to adversarial attacks.
If we conduct adversarial attacks on the samples $x_i$ before evaluating using (\ref{eqn:std}), the result is:
\begin{equation}
    E_{\text{attack}} \approx \frac{1}{N} \sum_{i=1}^N \max_{\delta_i \in \Delta} L(f(x_i + \delta_i), y(x_i))
    \label{eqn:aa}
\end{equation}

There are at least two key differences between adversarial attack (AA) and adversarial examination (AE).
First, AE deals with the underlying form whereas AA deals with the surface form. As a result, the variations considered by AE on the surface form is much larger than $\Delta$.
Second and more importantly, in AA there is a ``canonical'' starting point $\delta_i = 0$, so essentially all attack algorithms perform gradient descent starting from $\delta_i = 0$.
As a result, the sequence $\delta_i^1, \delta_i^2, \hdots$ is usually very local.
In AE, there is typically no ``canonical'' choice for $s_i$, so AE is forced to start with the entire space $\mathcal{S}$ instead of a specific point. 
As a result, the sequence $s_i^1, s_i^2, \hdots$ exhibits large variations.
This global picture reveals different properties of $f$ than AA, and is potentially more useful in representing the weakness (and/or strength) of the model. 

\paragraph{Availability of Underlying Representation}

While data in the surface form are straightforward to collect, there may be concerns whether it is always easy to obtain the underlying representation $z$ considered by the adversarial examiner. 
We argue that this problem has been greatly alleviated in recent years, with either rich annotation on real data \cite{xiang2014beyond,krishna2017visual}, or increasing use of synthetic data \cite{chang2015shapenet,johnson2017clevr}, or a hybrid between real and synthetic \cite{gaidon2016virtual,hudson2019gqa}.
In addition, in related topics such as image caption evaluation, the advantage of evaluating the underlying representation (e.g. SPICE \cite{anderson2016spice}) over the surface representation (e.g. BLEU \cite{papineni2002bleu}) has been demonstrated.

\subsection{Model Choices for Examiner}

We describe two classes of models as the \texttt{examiner} in Algorithm~\ref{alg:ae}. 
One is based on Reinforcement Learning (Section~\ref{sec:rl}), and the other is based on Bayesian Optimization (Section~\ref{sec:bo}). 
We conclude by discussing how these two classes of models are complementary (Section~\ref{sec:complementary}), which shows the general applicability of our adversarial examiner framework. 

\subsubsection{Reinforcement Learning as Examiner}
\label{sec:rl}

Let space $\mathcal{S}$ be the Cartesian product of $C$ factors $\mathcal{S} = \Psi^1 \times \Psi^2 \times \cdots \times \Psi^C$, where $\Psi^i$ represents the parameter range of the $i$-th factor. 
Therefore, the candidate $s_i^t$ is composed of $\psi_{(i,t)}^1, \psi_{(i,t)}^2, \hdots, \psi_{(i,t)}^C$, where $\psi_{(i,t)}^c \in \Psi^c$.
The probability of generating $s_i^t$ is $P(s_i^t) = \prod_{c=1}^C P(\psi_{(i,t)}^c | \psi_{(i,t)}^{c-1:1})$. 
We use a LSTM \cite{hochreiter1997long} to parameterize these conditional probabilities. 
Concretely, in the first LSTM step, the initial hidden state $h^1$ is followed by a fully connected layer and softmax to represent $P(\psi_{(i,t)}^1)$. 
We then draw a sample according to this distribution, which is fed into the second LSTM step. 
The updated hidden state $h^2$ is followed by a fully connected layer and softmax to represent $P(\psi_{(i,t)}^2 | \psi_{(i,t)}^1)$. 
This process is repeated until we have drawn a complete sample of $C$ steps. 

To train this LSTM using reinforcement learning, we define the reward signal $R$ to be $L(f(g(z_i, s_i^t)), y(z_i))$, and optimize the weights $\theta$ using policy gradient \cite{williams1992simple}:
\begin{equation}
    \nabla_\theta \mathbb{E}_{P(s_i^t; \theta)} [R] \approx \frac{1}{B} \sum_{b=1}^B \sum_{c=1}^C \nabla_\theta \log P(\psi_{(i,t)}^c | \psi_{(i,t)}^{c-1:1}) R_b
\end{equation}
where $B$ is the batch size to reduce high variance.

\subsubsection{Bayesian Optimization as Examiner} 
\label{sec:bo}

We use Gaussian Process (GP) as the model behind Bayesian optimization. 
The value that the GP aims to maximize is $L(f(g(z_i, s_i^t)), y(z_i))$ from Algorithm~\ref{alg:ae}.
The example generated by our examiner is equivalent to the point proposed by the acquisition function $a: \mathcal{S} \rightarrow \mathbb{R}^+$. 
We use Gaussian Process upper confidence bound (UCB) \cite{srinivas2009gaussian} to construct our acquisition function.

Let $W$ be the set of points that induce the posterior multivariate Gaussian distribution, which is initially empty.
For each iteration $t=1, 2, \hdots, T$, we select the next candidate by:
\begin{equation}
    s_i^t = \argmax_{s \in \mathcal{S}} a(s)
\end{equation}
where $a$ is the acquisition function induced by the current $W$. 
We then construct its surface form and test on the model. 
This newly observed point $(s_i^t, L(f(g(z_i, s_i^t)), y(z_i)))$ is then added to set $W$.
In the next iteration, the examiner will induce a new posterior multivariate Gaussian distribution based on the updated set $W$. 
By the end of adversarial examination, the candidates $\{ s_i^t \in \mathcal{S}\}_{t=1}^T$ are the points that induce the most up-to-date posterior multivariate Gaussian distribution on $\mathcal{S}$.

\subsubsection{How the Two Examiners Are Complementary}
\label{sec:complementary}

\paragraph{Discrete vs. Continuous} 
In RL, the choices of parameters in $\Psi$ are discrete as we use softmax to select amongst them. 
As for BO, the ranges of parameters are typically continuous as the underlying assumption is a multivariate Gaussian distribution.

\paragraph{Maintaining Sampling Distribution on $\mathcal{S}$ vs. Maintaining Function Value on $\mathcal{S}$} 
The LSTM model within RL captures the sampling probability $P$ on $\mathcal{S}$, without explicitly recording the function value $L$. 
On the other hand, the GP model within BO relies on the function values (set $W$) to fit the acquisition function. 

\paragraph{Longer Iteration Regime vs. Shorter Iteration Regime}
BO can be quite sample efficient, and is potentially able to successfully attack the model within a limited number of steps.
However, it becomes costly when the iteration gets long. 
On the other hand, policy-based RL can be inefficient, but the computation cost does not go up with more iterations. 

\section{Related Work}

We argue that the average case performance does not always reflect human's level of trust in a model. 
The primary design philosophy behind adversarial examiner is to place more emphasis on the worst case performance, which especially matters in sensitive domains. 
Indeed, medical imaging papers typically report the worst performance on the test set in addition to the average performance. 
The idea of paying more attention to the worst cases was recently discussed in \citet{yuille2018deep}. 

As suggested by its name, adversarial examiner is related to finding adversarial examples \cite{szegedy2013intriguing}. 
They were originally found by doing backpropagation and gradient ascent on the input image.
Later researchers started to consider the more challenging but more general scenario, where the neural network weights are not known to the attacker. 
This is called black box adversarial attack, where existing methods \cite{narodytska2016simple,chen2017zoo,ilyas2018black} still rely on estimating the gradient. 
Essentially, black box adversarial attack belongs to derivative-free optimization, which includes many other families of methods than gradient approximation. 
To the best of our knowledge, ours is the first work that brings Reinforcement Learning (RL) and Bayesian Optimization (BO) to black box adversarial attack. 
Our RL setup is inspired by work on Neural Architecture Search \cite{zoph2016neural}, whereas our BO setup is inspired by work on hyperparameter selection \cite{snoek2012practical}. 
Again, both problems are typical derivative-free optimization. 

Recently, researchers have been rethinking the concept of adversarial examples \cite{gilmer2018motivating} and generalizing them beyond modifying pixel values. 
\cite{engstrom2017rotation,pei2017towards,azulay2018deep} showed that deep networks can be attacked simply by 2D rotation, translation, or brightness change. 
\cite{kurakin2016adversarial,athalye2018synthesizing,eykholt2017robust} demonstrated the possibility of generating physical adversarial examples in the real world. 
\cite{gu2019using} used the minute transformations across video frames to study adversarial robustness.
\cite{brown2018unrestricted} introduced the concept of unrestricted adversarial examples, and \cite{song2018constructing} described a method to generate unrestricted adversarial examples using a differentiable neural network. 
More related to our work, \cite{zeng2017adversarial,yang2018realistic,alcorn2018strike,kanezaki2018rotationnet} incorporated rendering into the visual recognition pipeline to study how 3D physical parameters may affect model prediction. 
However, they were all concerned with creating adversarial examples starting from a specific configuration, instead of considering the global space. 

Finally, there are a couple of works that used synthetic data to iteratively improve the model, whether in the training stage or the inference stage. 
\citet{chen2018sampleahead} proposed to increase the sampling probability in the underperforming region. 
\citet{misra2018learning} used an agent to learn to select questions in synthetic visual question answering. 
\citet{yang2019embodied} learned to navigate in a synthetic scene and select better views for visual recognition.
\citet{kar2019meta} learned the policy to generate synthetic scenes, whose goal is to best facilitate downstream tasks. 
Our work is different, in that we focus on a new paradigm of model evaluation, which systematically explores the input data space to identify model weaknesses.

\section{Experiments}

\subsection{Implementation Details}

\renewcommand{\colwidth}{1.5cm}

\begin{table*}[t]
    \centering
\begin{tabular}{|C{1.5cm}|C{\colwidth}|C{\colwidth}|C{\colwidth}|C{\colwidth}|C{\colwidth}|}
\hline
Model & Examiner & $T=0$ & $T=100$ & $T=300$ & $T=500$ \\
\hline\hline
\multirow{2}{*}{AlexNet} & RL & $63.98\%$    & $65.91\%$    & $18.92\%$    & $2.27\%$     \\
 \cline{2-6}
                         & BO & $60.05\%$    & $43.58\%$    & $29.98\%$    & $25.43\%$     \\
\hline\hline
\multirow{2}{*}{ResNet34}  & RL & $69.03\%$     & $68.58\%$     & $38.86\%$    & $13.13\%$     \\
\cline{2-6}
                         & BO & $64.19\%$    & $54.89\%$     & $48.07\%$    & $45.55\%$     \\
\hline
\end{tabular}
    \caption{Average model performance under adversarial examination. 
    We report $2 \times 2 \times 4 = 16$ settings under various model, examiner, and number of iterations combinations. 
    The values are average post-softmax probability on the true class. }
    \label{tab:main}
\end{table*}

\begin{figure*}[t]
    \begin{subfigure}[b]{0.5\textwidth}
    \centering
    \includegraphics[width=0.9\linewidth]{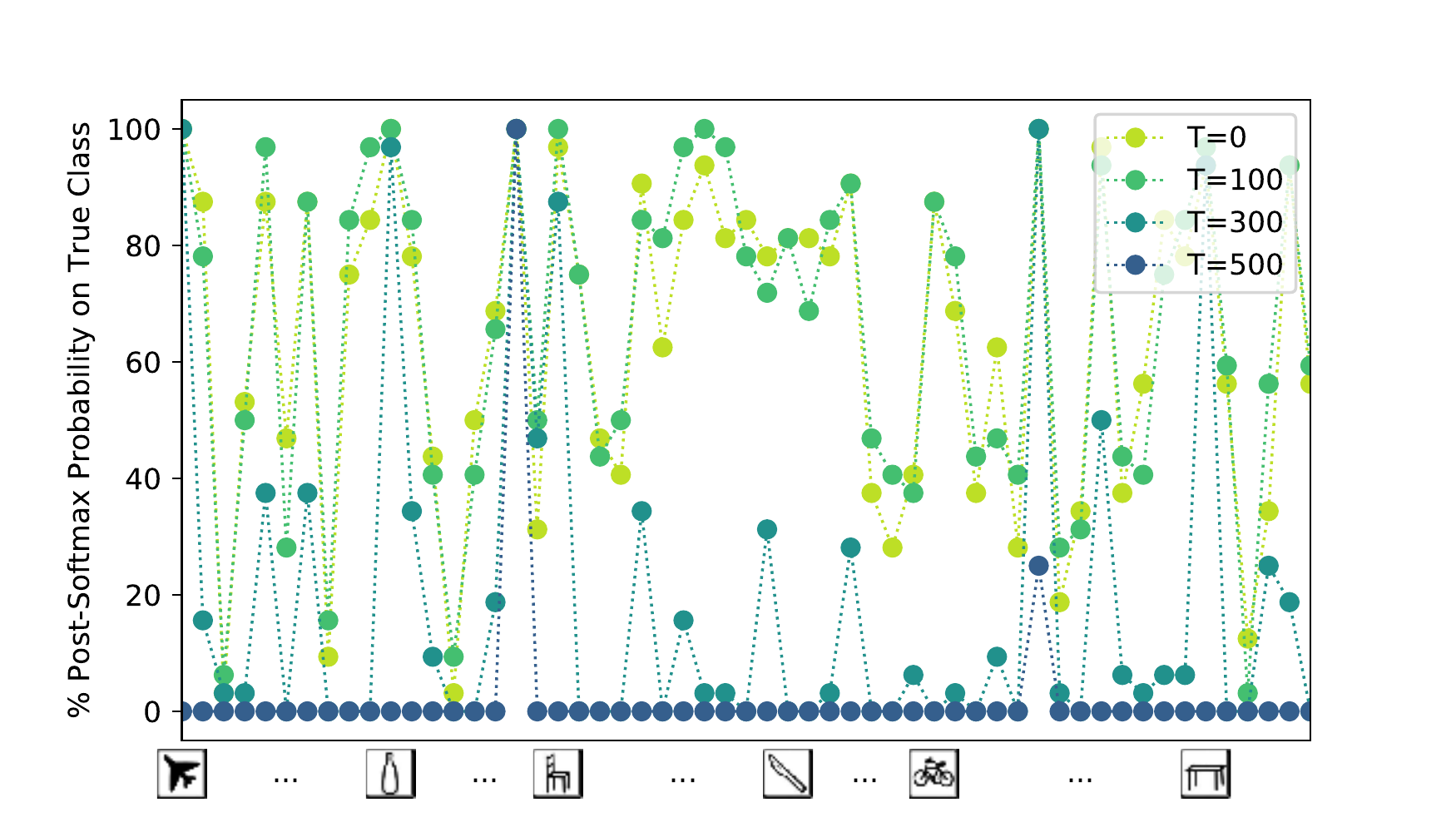}
    \caption{RL Examiner on AlexNet}
    \end{subfigure}
    \begin{subfigure}[b]{0.5\textwidth}
    \centering
    \includegraphics[width=0.9\linewidth]{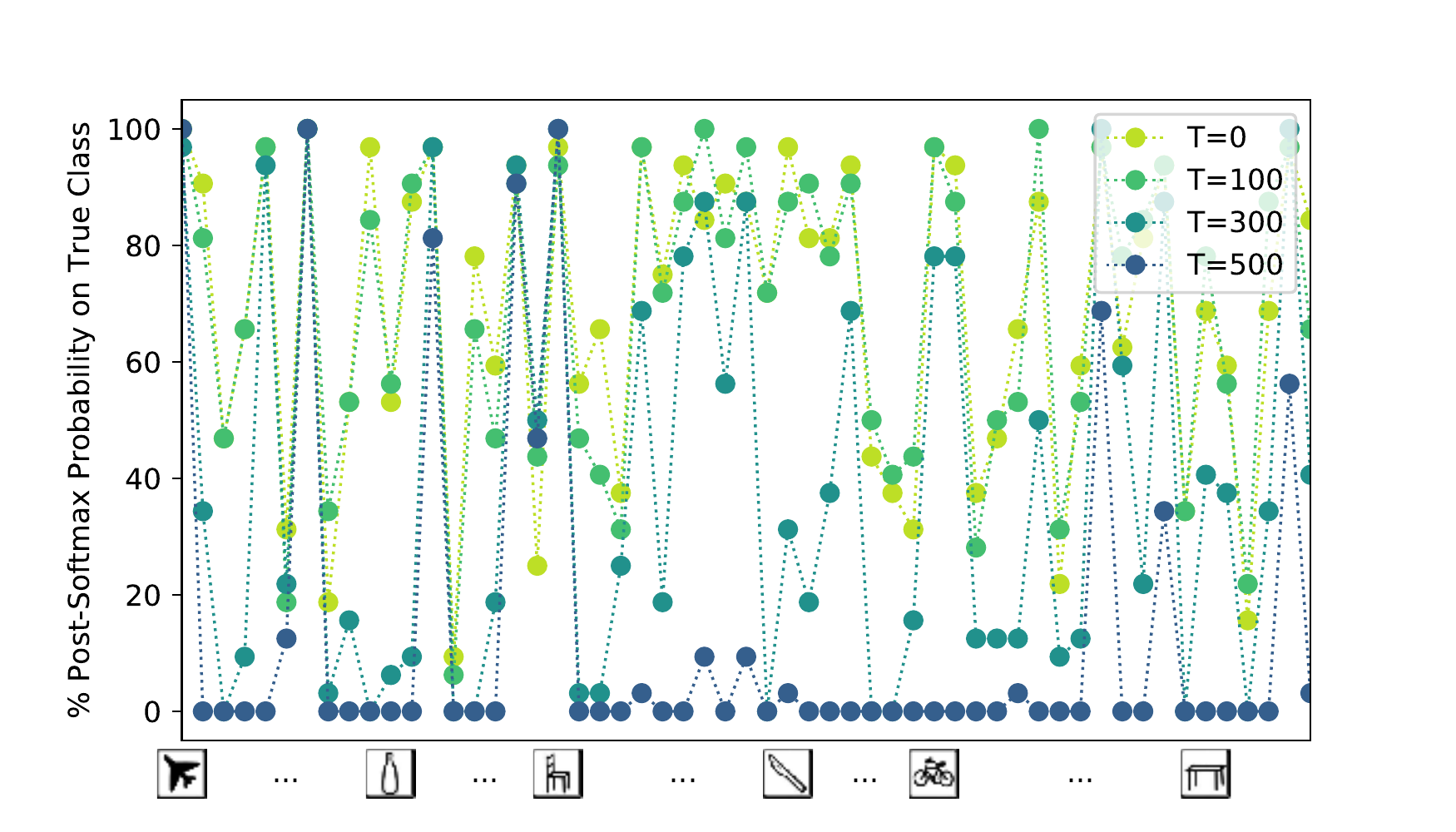}
    \caption{RL Examiner on ResNet34}
    \end{subfigure}
    \begin{subfigure}[b]{0.5\textwidth}
    \centering
    \includegraphics[width=0.9\linewidth]{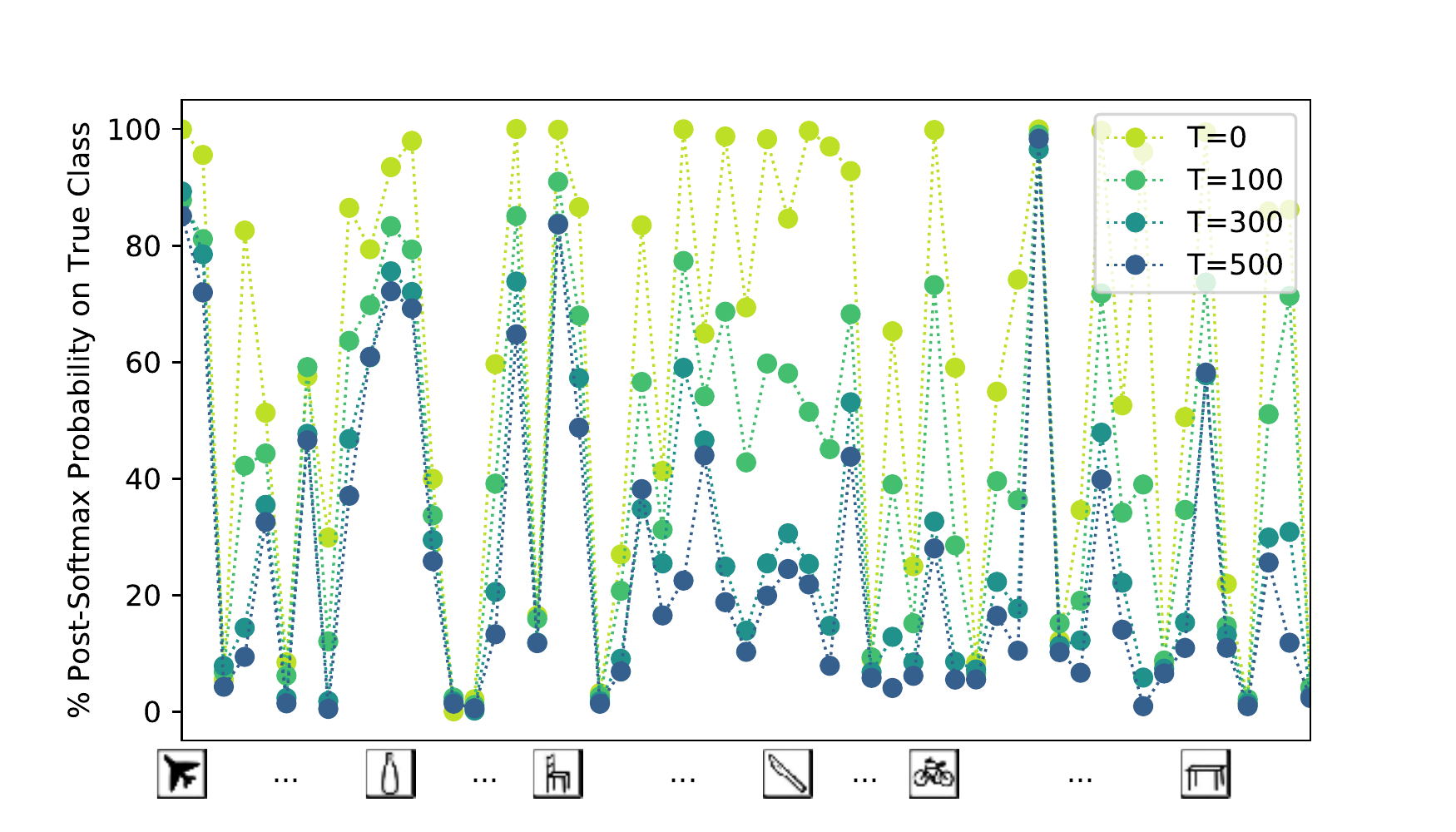}
    \caption{BO Examiner on AlexNet}
    \end{subfigure}
    \begin{subfigure}[b]{0.5\textwidth}
    \centering
    \includegraphics[width=0.9\linewidth]{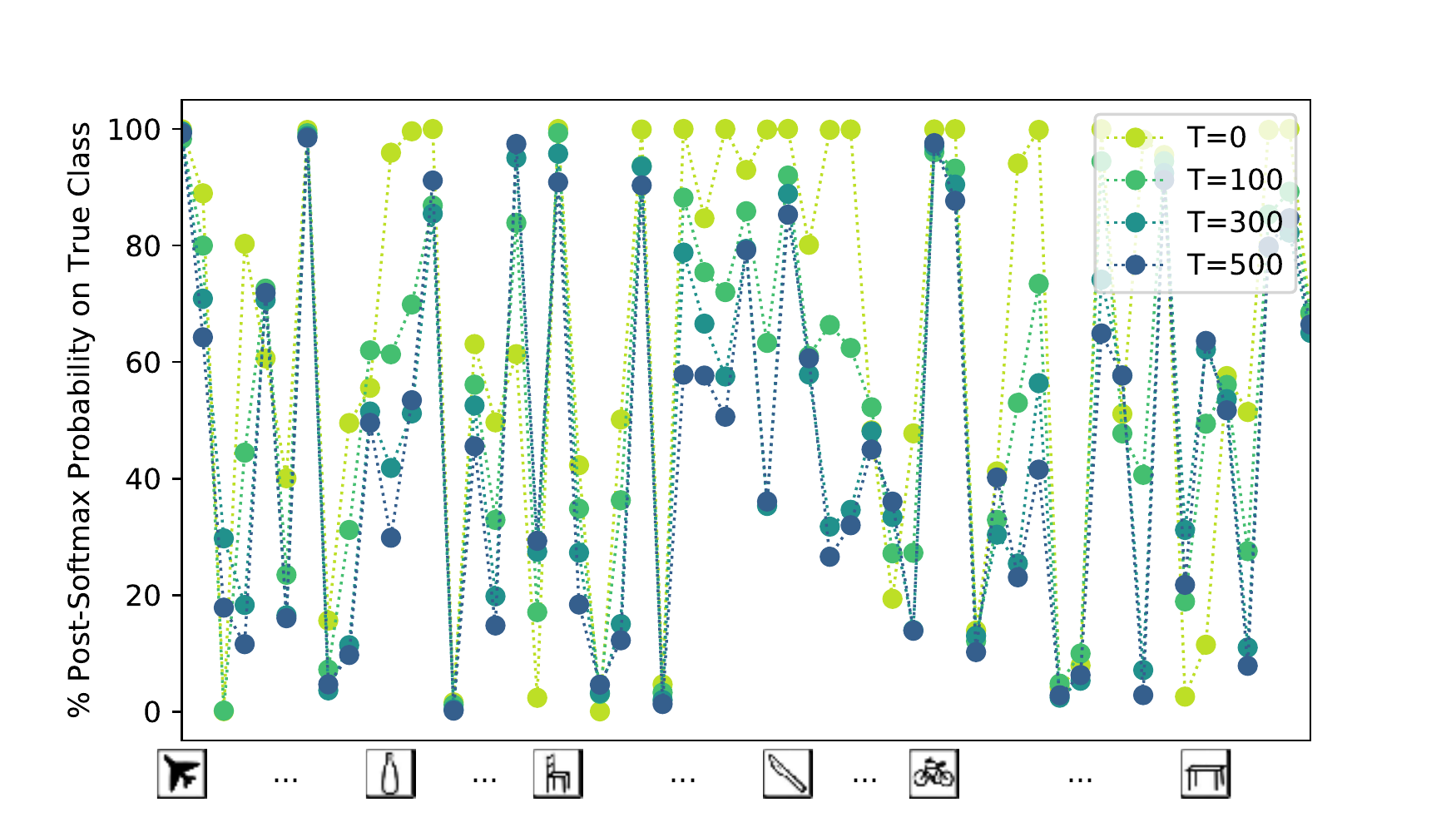}
    \caption{BO Examiner on ResNet34}
    \end{subfigure}
    \caption{
    Per-class model performance under adversarial examination. 
    The four plots correspond to AlexNet (left) vs. ResNet34 (right), RL (upper) vs. BO (lower). 
    Horizontal axis is object category. 
    AlexNet is more vulnerable than ResNet34, and the RL examiner seems more strict than BO examiner under the same $T$.}
    \label{fig:examine}
\end{figure*}

\begin{figure*}[t]
    \centering
    \begin{subfigure}[b]{0.24\textwidth}
    \includegraphics[width=\linewidth]{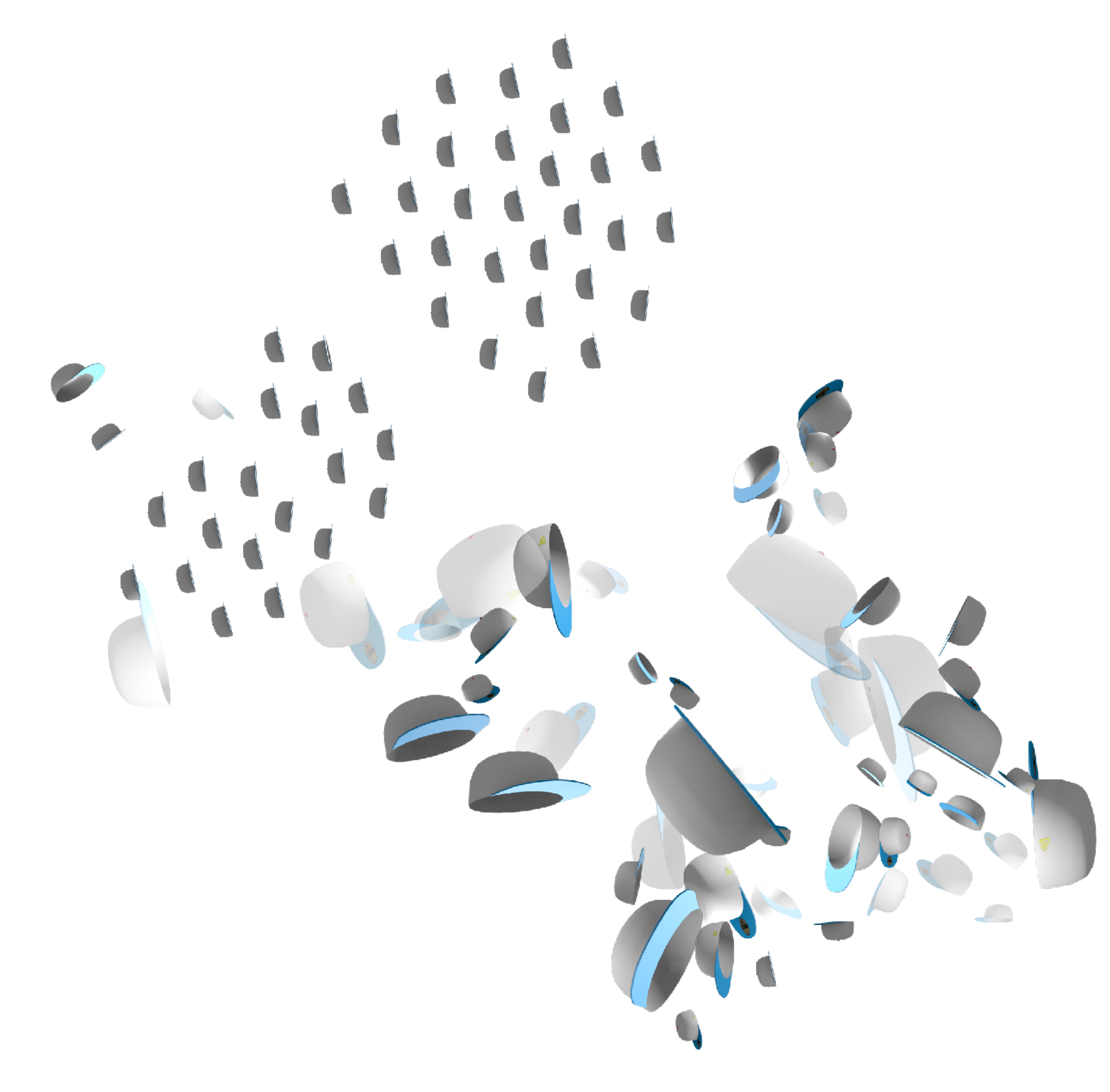}
    \caption{RL on AlexNet}
    \end{subfigure}
    \begin{subfigure}[b]{0.24\textwidth}
    \includegraphics[width=\linewidth]{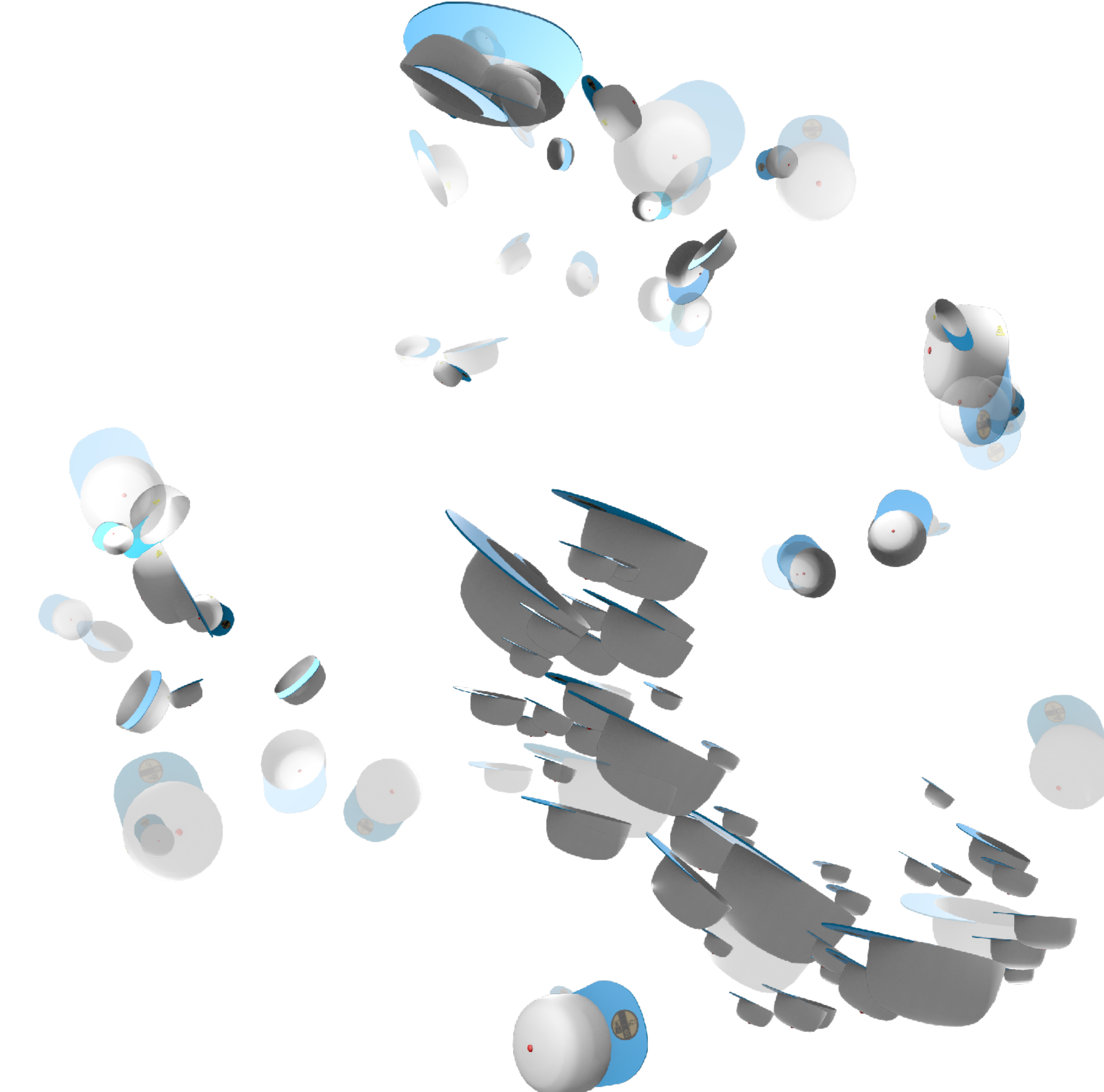}
    \caption{BO on AlexNet}
    \end{subfigure}
    \begin{subfigure}[b]{0.24\textwidth}
    \includegraphics[width=\linewidth]{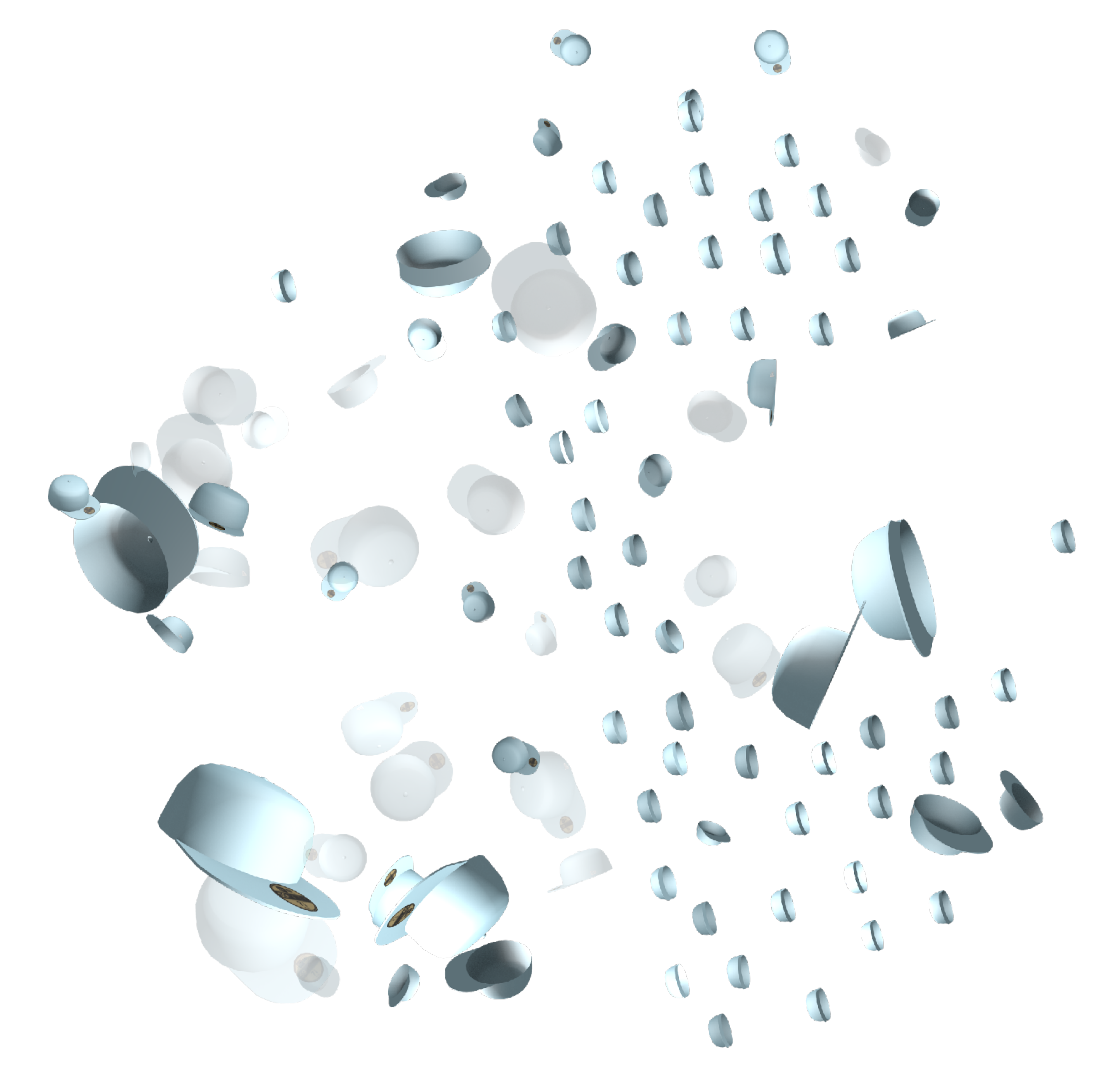}
    \caption{RL on ResNet34}
    \end{subfigure}
    \begin{subfigure}[b]{0.24\textwidth}
    \includegraphics[width=\linewidth]{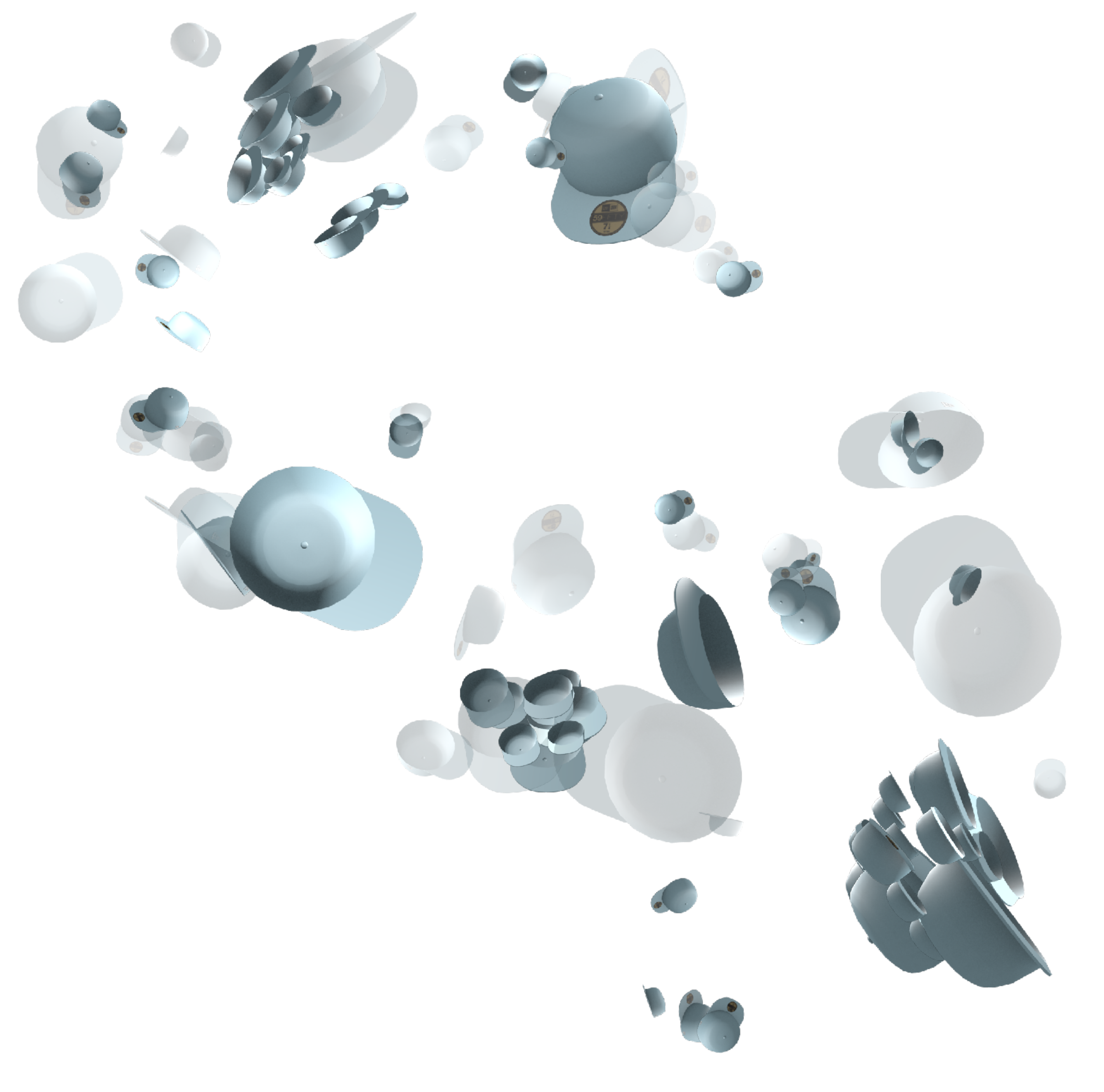}
    \caption{BO on ResNet34}
    \end{subfigure}
    \caption{t-SNE visualization of \textit{cap} examples under different examiners and target models. There are $100$ examples in each subfigure: $50$ randomly sampled ones, and $50$ from the very end of adversarial examination. Transparent cap means correctly classified. Otherwise, incorrectly classified. }
    \label{fig:tsne}
\end{figure*}

We conduct experiments on visual recognition of objects in the ShapeNet dataset \cite{chang2015shapenet}, which contains $55$ classes and $51190$ instances. 
Here, the underlying representation $z$ would be the 3D object, $s$ would be parameters required in rendering (we consider a total of $C = 12$ factors, listed in Tab.~\ref{tab:factors}), and surface form $x$ would be the 2D image after rendering. 
We use the Blender software for rendering. 

Our target models are ResNet34 \cite{he2016deep} and AlexNet \cite{krizhevsky2012imagenet} trained on the 2D image surface form $x$. The ResNet34 model is trained with learning rate of $0.005$, and AlexNet model with $0.001$, both with Adam optimizer \cite{kingma2014adam} for $40$ epochs.

During training, we randomly select a value for each of the $12$ factors, and render $m=10$ images per 3D object. During adversarial examination, we assume that the examiners have no control over the location of the sun, so we randomly choose the rotation angles for the sun and fix them. The remaining $9$ factors will be available to and explored by the examiners. 
For each class, we choose one 3D object in the validation set that has the highest post-softmax probability on the true class. 

In RL, the $9$ continuous factors will be discretized to $100$ choices evenly distributed on their respective ranges. When sampling the factors, the choice made will be mapped to its corresponding embedding space (embedding size $30$), then fed into LSTM (hidden state size $30$). We set the learning rate to $0.001$ and batch size to $32$, and use Adam optimizer \cite{kingma2014adam} to update model parameters. In BO, our implementation is based on the \texttt{Bayesian Optimization} package\footnote{\url{https://github.com/fmfn/BayesianOptimization}} and we allow $2$ random examples at the very beginning. Both examiners operate for $T = 500$ iterations, and $L$ is the negative post-softmax probability on the true class. 

\renewcommand{\colwidth}{1.2cm}
\begin{table}[t]
    \centering
\begin{tabular}{|C{\colwidth}|C{\colwidth}|C{\colwidth}|C{\colwidth}|C{\colwidth}|}
\hline
   & $m=10$ & $m=5$ & $m=2$ & $m=1$ \\
\hline\hline
RL & $63.81\%$    & $57.43\%$    &  $35.05\%$   & $18.92\%$     \\
\hline
BO & $49.79\%$    & $43.06\%$    & $22.19\%$    & $10.92\%$      \\
\hline
\end{tabular}
    \caption{Average model performance under adversarial examination for varying $m$: number of training images per instance. We report $m = 1, 2, 5, 10$ with iterations $T = 200$.}
    \label{tab:diff_training}
\end{table}

\begin{figure}[t]
    \centering
    \includegraphics[width=0.6\linewidth]{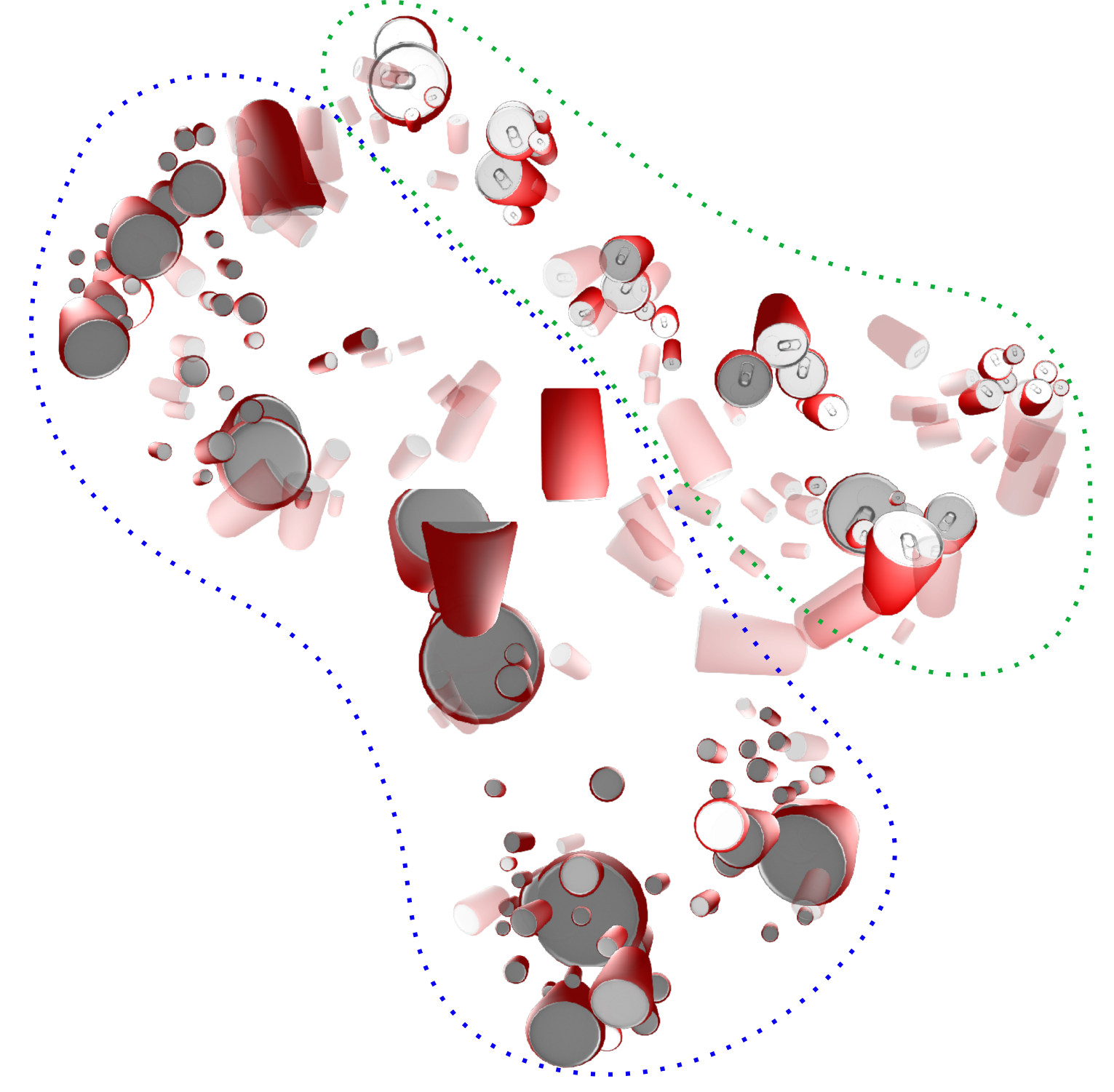}
    \caption{t-SNE visualization of \textit{can} examples. $100$ are randomly sampled, and $100$ are from RL examiner $T = 250$. Transparent can means correctly classified. Otherwise, incorrectly classified. In this experiment, training data for ResNet34 did not have top and bottom viewpoints. Adversarial examiner is able to identify these two artificial weaknesses (circled in green and blue). }
    \label{fig:hole_tsne}
\end{figure}

\subsection{Evaluating Model Performance with Adversarial Examiners}

The evaluation results by our adversarial examiners are shown in Tab.~\ref{tab:main}. 
For both AlexNet and ResNet34, we tried both RL and BO examiners under different number of iterations $T$. 
The $T = 0$ column corresponds to the standard evaluation protocol, which on average records more than $60\%$ probability on the true class. 
After a few hundred iterations, both examiners can find the weakness of the model successfully. In addition, longer examination results in lower performance, which is expected. 
But their characteristics are not entirely the same. 
BO can find weaknesses relatively faster (larger gap from $T = 0$ to $T = 100$), but eventually ($T = 300$ or $T = 500$) RL becomes more committed to the weaknesses and gives a harsher evaluation. 

Fig.~\ref{fig:examine} shows the per-class post-softmax probability on the true class. We found that the weaknesses of some classes are easier to find than others. For instance, \textit{barrel} and \textit{bowl} are relatively vulnerable since they look very similar (generic round shape) when viewing from bottom-up. 
Classes such as \textit{airplane}, which contain discriminative visual cues at all angles, are more robust under adversarial examination.

Fig.~\ref{fig:tsne} uses t-SNE \cite{maaten2008visualizing} to visualize $50$ randomly sampled examples and $50$ examples selected at the end of adversarial examination of the \textit{cap} class. 
Correctly classified examples are transparent, and incorrectly classified ones are opaque. 
This visualization serves to show the global distribution of model weakness on the space $\mathcal{S}$. 
We can observe clusters among the incorrectly classified examples, e.g. due to viewpoint or lighting. 

\subsection{Examining Models Trained with Less Data}

Intuitively, neural networks trained with less data should be less robust against adversarial examiners. 
In this experiment, we put this hypothesis to test. 
Specifically, we vary $m$, the number of images rendered for each 3D object in the training set. 
We train ResNet34 models with $m = 1, 2, 5, 10$ (recall that $m = 10$ is the default) using the same number of epochs. 

Tab.~\ref{tab:diff_training} summarizes the results. 
Under the same iterations $T$, both RL and BO examiners report decreased performance as $m$ decreases. 
In addition to validating the adversarial examiner framework, this also confirms the importance of the amount and diversity of training data.

\begin{figure*}[t]
    \centering
    \includegraphics[width=0.8\textwidth]{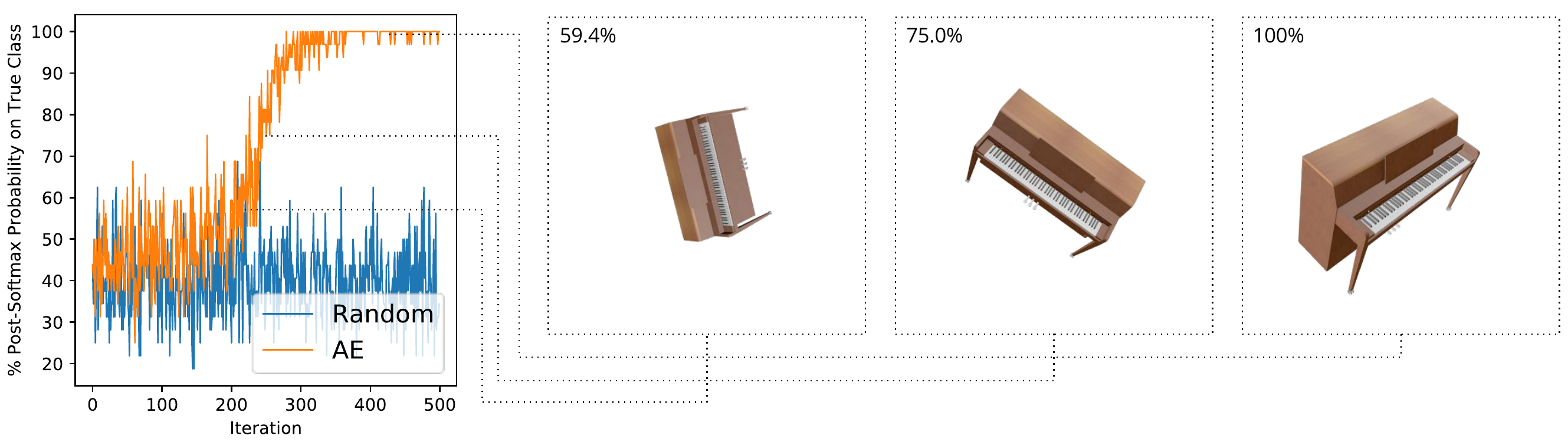}
    \caption{Identifying model strength by negating $L$. For this \textit{piano} instance, the examiner eventually finds a condition under which the model can correctly classify with $100\%$ confidence. }
    \label{fig:inverse}
\end{figure*}

\subsection{Evaluating Model with Artificial Weaknesses}

To further study the properties of adversarial examiner, we consider an artificial scenario, where we deliberately put in obvious weaknesses. 
We are interested to see to what extent can the adversarial examiner recover these obvious weaknesses. 
Specifically, during training, we limit the viewpoint elevation $U_v$ to $[-30^\circ, +30^\circ]$ instead of $[-90^\circ, +90^\circ]$. 
This will create two obvious weaknesses: viewing objects from the top and from the bottom. 
Note that the examiner is still allowed to change all $9$ factors. 

Fig.~\ref{fig:hole_tsne} shows the t-SNE visualization of $200$ \textit{can} examples. 
$100$ are randomly sampled, and $100$ are from $T = 250$ during adversarial examination. 
Among the incorrectly classified examples (visualized as opaque), there are two large regions, with the bottom left representing viewing from the bottom, and the top right representing viewing from the top. 
This observation further confirms the capability of adversarial examiners. 

\subsection{Changing the Order of Factors}

\begin{figure}[b!]
    \centering
    \begin{subfigure}[b]{0.49\linewidth}
    \includegraphics[width=0.95\linewidth]{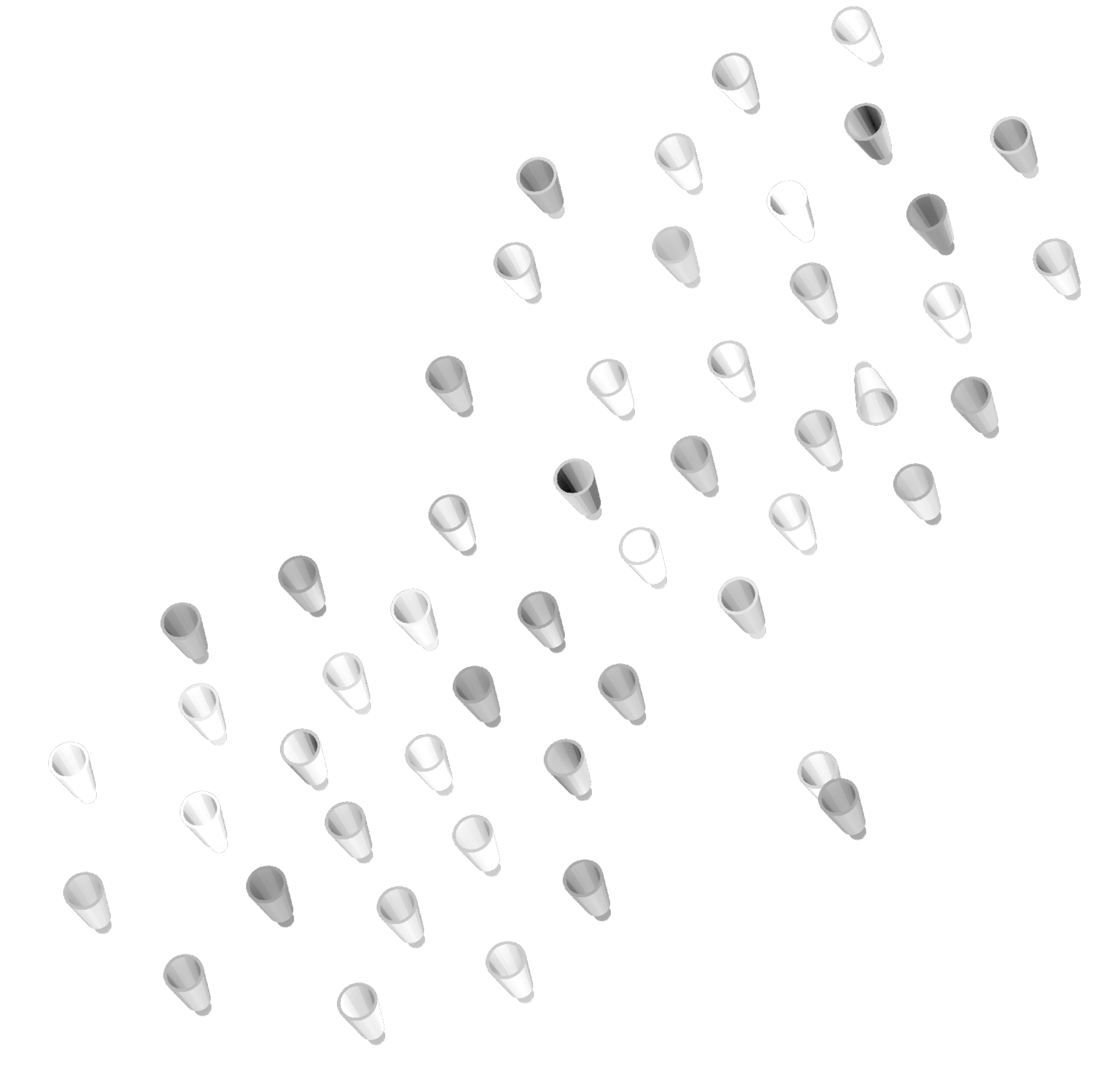}
    \caption{$\Gamma_{l}, \Gamma_{o}, r_{l}, A_{l}, U_{l}, r_{v}, A_{v}, U_{v}, \theta_{v}$}
    \end{subfigure}
    \begin{subfigure}[b]{0.49\linewidth}
    \includegraphics[width=0.95\linewidth]{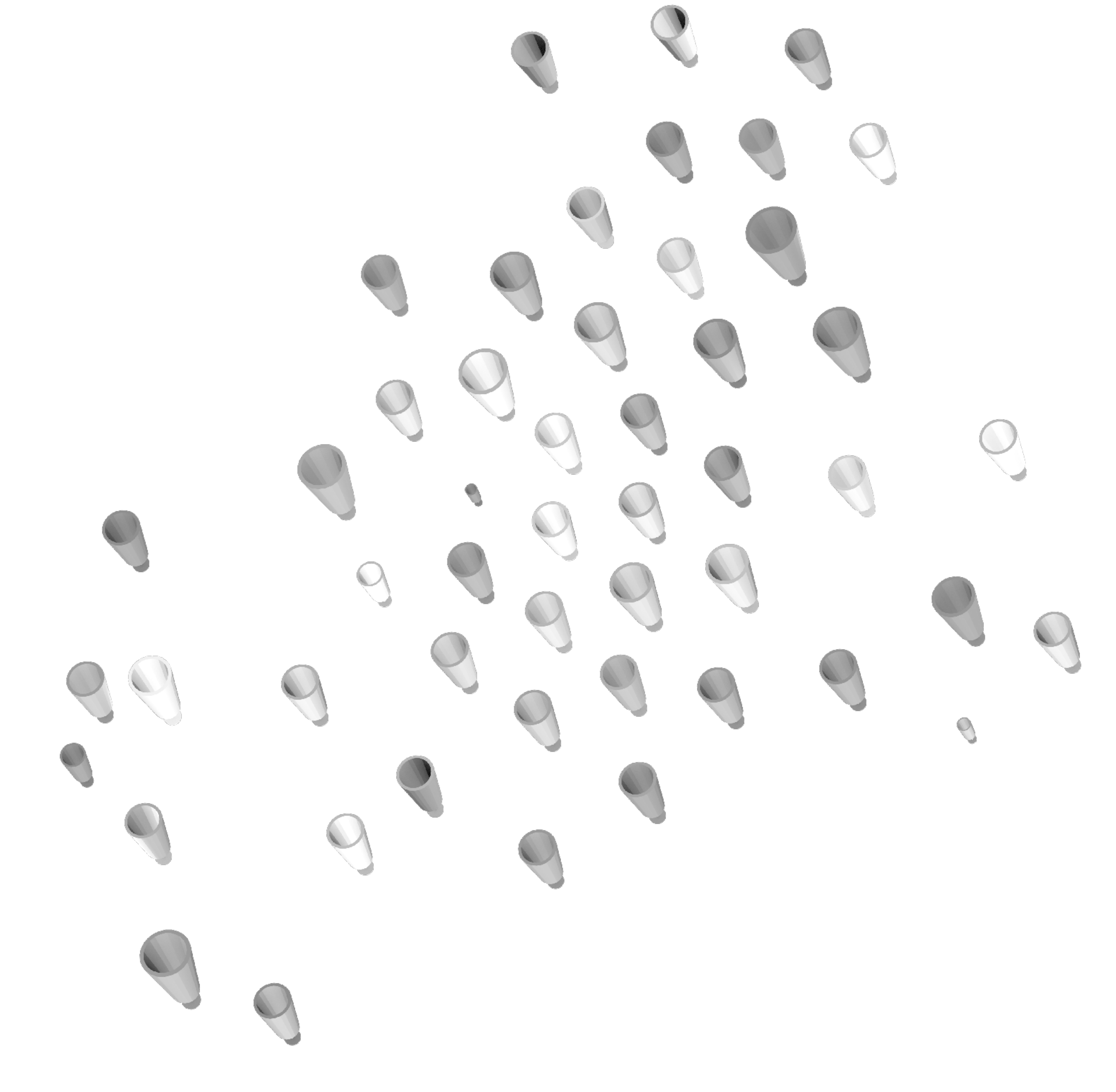}
    \caption{$ r_{v}, A_{v}, U_{v}, \theta_{v}, \Gamma_{l}, \Gamma_{o}, r_{l}, A_{l}, U_{l}$}
    \end{subfigure}
    \caption{The last $50$ \textit{lamp} examples given by two RL examiners with different order of factors. }
    \label{fig:order}
\end{figure}

As described in Section~\ref{sec:rl}, the RL examiner samples parameters in a sequential manner. In this experiment, we investigate whether changing the sampling order of factors will make a difference in adversarial examination.

The default RL examiner samples in the order $\Gamma_{l}, \Gamma_{o}, r_{l}, A_{l}, U_{l}, r_{v}, A_{v}, U_{v}, \theta_{v}$. 
We prepare a second RL examiner with exactly the same setup, but samples in the order $ r_{v}, A_{v}, U_{v}, \theta_{v}, \Gamma_{l}, \Gamma_{o}, r_{l}, A_{l}, U_{l}$. 
Fig.~\ref{fig:order} shows the t-SNE visualization of the last $50$ examples given by the two RL examiners ($T = 500$). 
Obviously they have converged to the same weaknesses, which is evidence that the RL examiner is not sensitive to the order. 

\subsection{Identifying Model Strength}

Finally, we show that instead of identifying model weakness, the same setup can also be used to identify model strength, simply by flipping the $\max$ in (\ref{eqn:ae}) to $\min$. 
The examiner, instead of trying to find the worst viewing condition of an object, now tries to find the best viewing condition. 

We show such a trial in Fig.~\ref{fig:inverse}. 
We indeed observe a curve where the post-softmax probability on the true class goes up throughout evaluation, which is symmetric to Fig.~\ref{fig:examine_examples}. 
The final image reveals what the model considers to be the easiest \textit{piano} to recognize. 

\section{Conclusion}

In this paper, we advocate for a new testing paradigm for machine learning models, where more emphasis is placed on the worst case instead of reporting the average case performance. 
To make this idea concrete, we propose \emph{adversarial examiner}, whose goal is to undermine the model's performance by discovering and concentrating on its weaknesses. 
The adversarial examiner will dynamically hand out the next testing data, based on the testing history so far. 
As a result, different models will now be evaluated using potentially different samples from the input data space. 
We argue that this notion can more effectively prevent hype, is closer in spirit to human interview/Turing test, and can be especially important in sensitive domains such as autonomous driving and medical applications. 

We conduct experiments on visual recognition of ShapeNet objects. 
Examiners based on Reinforcement Learning and Bayesian Optimization both demonstrated effectiveness in identifying the weaknesses of the target model. 
However, our adversarial examination framework is much more general than classification of rendered 3D objects.
In the future, we hope to extend to other domains, including image generation, language generation, and pose estimation in robotics. 
Using the examples generated during adversarial examination to improve training is another potentially fruitful venue. 

\section*{Acknowledgments}

This work is supported by IARPA via DOI/IBC contract No. D17PC00342. CL is supported by Google PhD Fellowship, and also acknowledges NVIDIA for awarding a GPU used in this research. 

\bibliographystyle{aaai}
\bibliography{AAAI-ShuM.8993.bib}

\end{document}